\documentclass{article}
\usepackage{ijcai25}

\usepackage{times}
\usepackage{soul}
\usepackage{url}
\usepackage[hidelinks]{hyperref}
\usepackage[utf8]{inputenc}
\usepackage[small]{caption}
\usepackage{graphicx}
\usepackage{amsmath}
\usepackage{amsthm}
\usepackage{booktabs}
\usepackage{algorithm}
\usepackage{algorithmic}
\usepackage[switch]{lineno}
\usepackage{amssymb}
\usepackage{makecell}
\usepackage{natbib}
\usepackage[hidelinks]{hyperref}

\newtheorem{theorem}{Theorem}
\newtheorem{proposition}[theorem]{Proposition}

\title{Evolvable Conditional Diffusion}

\author{
    Author Name
    \affiliations
    Affiliation
    \emails
    email@example.com
}

\author{
Zhao Wei$^{1,2}$
\and
Chin Chun Ooi$^{1,2}$\thanks{Corresponding author}\and
Abhishek Gupta$^3$\and
Jian Cheng Wong$^2$\and
Pao-Hsiung Chiu$^2$\and
Sheares Xue Wen Toh$^2$\And
Yew-Soon Ong$^{1,4}$\\
\affiliations
$^1$Centre for Frontier AI Research, Agency for Science, Technology and Research, Singapore\\
$^2$Institute of High Performance Computing, Agency for Science, Technology and Research, Singapore\\
$^3$School of Mechanical Sciences, Indian Institute of Technology, Goa, India\\
$^4$College of Computing and Data Science, Nanyang Technological University, Singapore\\
\emails
\{wei\_zhao, ooicc\}@cfar.a-star.edu.sg,
abhishekgupta@iitgoa.ac.in, \{wongj, chiuph, sheares\_toh\}@ihpc.a-star.edu.sg, asysong@ntu.edu.sg
}

\begin{document}

\maketitle

\begin{abstract}
    This paper presents an evolvable conditional diffusion method such that black-box, non-differentiable multi-physics models, as are common in domains like computational fluid dynamics and electromagnetics, can be effectively used for guiding the generative process to facilitate autonomous scientific discovery. We formulate the guidance as an optimization problem where one optimizes for a desired fitness function through updates to the descriptive statistic for the denoising distribution, and derive an evolution-guided approach from first principles through the lens of probabilistic evolution. Interestingly, the final derived update algorithm is analogous to the update as per common gradient-based guided diffusion models, but without ever having to compute any derivatives. We validate our proposed evolvable diffusion algorithm in two AI for Science scenarios: the automated design of fluidic topology and meta-surface. Results demonstrate that this method effectively generates designs that better satisfy specific optimization objectives without reliance on differentiable proxies, providing an effective means of guidance-based diffusion that can capitalize on the wealth of black-box, non-differentiable multi-physics numerical models common across Science.
\end{abstract}

\vspace{0mm}
\section{Introduction}

Diffusion models have emerged as a powerful class of deep generative models, demonstrating remarkable performance in a variety of domains such as image synthesis \cite{rombach2022high}, audio generation \cite{kong2021diffwave}, biological sequence generation \cite{pmlr-v202-avdeyev23a}, and engineering design problems \cite{liu2024uncertainty,xulooks}. 

While the original unconditional diffusion model generates samples randomly, conditional diffusion models are a common variant that have been proposed to enable generation with a bias towards specific users' requirements (e.g., aesthetics and/or specific engineering performance criteria) \cite{yang2023diffusion}. For example, molecular linker design leverages conditioning on target protein pockets to advance drug discovery \cite{igashov2024equivariant}. However, conditional diffusion models require paired data-sets to be available \textit{a priori} for training, which can be difficult to obtain in large quantities.



In contrast, guided diffusion models and their variants \cite{dhariwal2021diffusion,bansal2023universal,maze2023diffusion,chen2024overview} have been proposed whereby the gradient of a regressor or classifier is used to steer the denoising trajectory such that samples with specific requirements are preferentially generated. Critically, guided diffusion models offer flexibility in conditional generation as they enable a pre-trained diffusion model to generate samples which satisfy any post-hoc requirement, a notable advantage as pre-trained diffusion models continue to proliferate and scale in the era of foundation models \cite{li2024derivative}. This has already led to several exciting advances in AI for Science, e.g., MatterGen \cite{zeni2025generative} demonstrates the ability of guided diffusion to directly generate novel, stable, synthesizable inorganic materials with specific target properties. The integration of such automated candidate generation strategies with autonomous laboratories of the future approach is key to achieving the vision of end-to-end autonomous scientific discovery \cite{szymanski2023autonomous}.

Although promising, practical applications of guided diffusion models still face certain challenges. One key issue is the need for models that can evaluate the desired performance to be differentiable in order for the gradient to be easily computed and incorporated during the denoising process. However, this precludes the use of many established multi-physics numerical models that have been developed for diverse scientific applications as the majority of such models are non-differentiable. These models are essentially black-boxes in that users typically do not have the capability to interact in an intrusive manner with the solver, e.g., molecular/protein descriptors \cite{gainza2020deciphering,ghiringhelli2015big} and physics-based simulators \cite{yuan2023physdiff}. Hence, this disconnect between the need for gradients and the limitations of current state-of-the-art high-fidelity solvers has been a major barrier to the applicability of performance-guided diffusion models in enabling major scientific transitions.

The methods of evolutionary computation offer a unique pathway to overcome this barrier. In contrast to local gradient-based guidance methods, the generative processes of evolution are inherently derivative-free and exploratory, driven by random genetic variation and guided by the principle of survival of the fittest. Critically, evolutionary computation's independence from gradient information makes it uniquely suited for exploiting the repertoire of existing black-box solvers established in many scientific domain, facilitating automated scientific discovery as described above while retaining compliance with existing known physics through evaluation and verification with state-of-the-art numerical solvers. Being a search-oriented approach, evolutionary computation is particularly effective for global exploration in complex, multi-modal landscapes, enabling it to tackle discrete, discontinuous, and high-dimensional problems as is common in many scientific domains, e.g., meta-surface design, a pressing challenge in communications and next-generation semiconductors. Moreover, its robustness in handling multiple guidance signals (fitness functions) and noisy environments reinforces its applicability to complex design challenges of today \cite{bali2020cognizant}. 

It is worth highlighting that the stochastic nature of evolutionary computation also promotes diversity and can unveil unexpected and innovative solutions, with prior literature demonstrating how this can lead to the emergence of novel biological structures \cite{miikkulainen2021biological}. Hence, the incorporation of in silico evolution in generative AI methods like diffusion models also engenders a similar potential to produce innovative, out-of-distribution solutions, making it an exciting engine for scientific discovery \cite{wong2024llm2fea,lyu2024covariance,lehman2020surprising}. Overall, evolutionary computation is a compelling alternative to traditional, gradient-based generative AI methods.

\begin{figure*}[t]
\centering
\vspace*{0mm}
\includegraphics[width=0.7\textwidth]{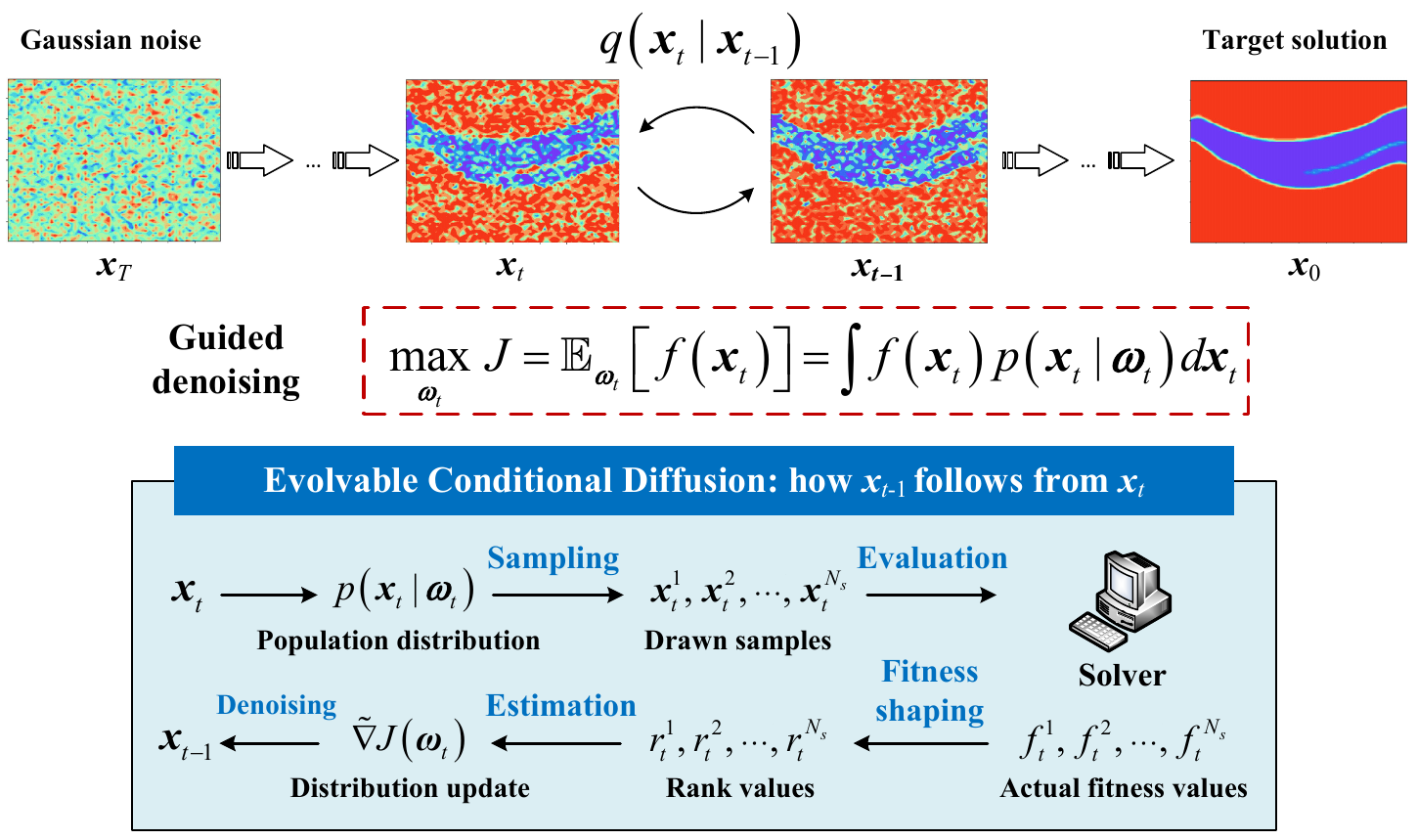}
\vspace*{-0mm}
\caption{Framework of the proposed evolvable conditional diffusion method. The entire process is framed through the lens of probabilistic evolvability, transitioning from $\boldsymbol{x}_T$ to $\boldsymbol{x}_0$ via a sequence of mutations \cite{valiant2009evolvability}. Notably, our derived update algorithm directly utilizes evolved samples and corresponding fitness values, eliminating the need for differentiable evaluation models during guidance and extending its applicability to non-differentiable solvers.}
\vspace*{-0mm}
\label{Fig_framework}
\end{figure*}

Given the gradient-free capabilities of evolutionary computation \cite{le2013evolution,sung2023neuroevolution}, we seek to develop a gradient-free approach to achieve guided generation through the lens of \textit{evolvability} \cite{valiant2009evolvability} of the diffusion model as shown in Figure \ref{Fig_framework}. 
This method enables generation of designs that preferentially align with desired objectives even if the evaluation tool (e.g., solver) is non-differentiable. 
The contributions are summarized as follows:

\begin{itemize}
    \item We propose an evolvable conditional diffusion method inspired by evolutionary algorithms. This approach treats the generation process as a black-box optimization problem, whereby the probabilistic distribution obtained from the pre-trained diffusion model is evolved to favor designs that maximize certain performance criteria. Notably, the derived update algorithm can be seen to be analogous to the update rule in conventional gradient-based guided diffusion under specific assumptions.
    \item The proposed method eliminates the need for differentiable models in the guided diffusion. Instead, the gradient of the fitness function is directly estimated from samples drawn from the evolved distribution, with corresponding fitness values evaluated using non-differentiable solvers.
    \item We successfully apply the proposed approach to two generative design problems as examples of how this method can be useful in the general area of AI for Science, i.e., the design of fluidic channel topology and frequency-selective meta-surface.
\end{itemize}

\vspace{0mm}
\section{Background and Related Works}
\subsection{Diffusion Models}

Diffusion models \cite{ho2020denoising,song2021maximum,song2019generative,song2020improved,song2020score} are a class of probabilistic generative models that learn to generate new samples by iteratively removing noise (denoising) from a random input. They have demonstrated advantages over other state-of-the-art methods like generative adversarial networks (GANs) across various tasks such as computer vision \cite{croitoru2023diffusion}, natural language processing \cite{li2022diffusion}, and topology design \cite{maze2023diffusion}.

Taking the famous denoising diffusion probabilistic model (DDPM) \cite{ho2020denoising} as an example, the training process consists of two main phases: a forward noising process and a reverse denoising process. In the forward noising process, data is progressively perturbed by adding Gaussian noise in small steps, eventually transforming the data into pure noise as per Eq. (\ref{diffusion process})

\vspace{0mm}
\begin{equation}
    q(\boldsymbol{x}_t | \boldsymbol{x}_{t-1}) = \mathcal{N}(\boldsymbol{x}_t; \; \sqrt{1 - {\beta}_t}\boldsymbol{x}_{t-1}, \; {\beta}_t\boldsymbol{I})
\label{diffusion process}
\end{equation}

\noindent where ${\beta}_t \in (0, 1)$ is a hyperparameter selected prior to model training. In contrast, the denoising process gradually removes the noise by running a learnable Markov chain to generate new data as illustrated in Eq. (\ref{denoising process})

\vspace{0mm}
\begin{equation}
    p_{\boldsymbol{\theta}}(\boldsymbol{x}_{t-1} | \boldsymbol{x}_t) = \mathcal{N}(\boldsymbol{x}_{t-1}; \; \boldsymbol{\mu}_{\boldsymbol{\theta}}(\boldsymbol{x}_t), \; \boldsymbol{\Sigma}_{\boldsymbol{\theta}}(\boldsymbol{x}_t))
\label{denoising process}
\end{equation}

\noindent where $\boldsymbol{\mu}_{\boldsymbol{\theta}}(\boldsymbol{x}_t)$ and $\boldsymbol{\Sigma}_{\boldsymbol{\theta}}(\boldsymbol{x}_t)$ are the mean and covariance of the denoising process, and are predicted through the trained neural network parameterized by $\boldsymbol{\theta}$. In some instances, the covariance  $\boldsymbol{\Sigma}_{\boldsymbol{\theta}}(\boldsymbol{x}_t)$ can be simplified as a constant \cite{ho2020denoising}. Once learned, new samples (e.g., images) can be generated by sampling from $\mathcal{N}(0, 1)$ and carrying out a denoising process according to Eq. (\ref{denoising process}).

\vspace{0mm}
\subsection{Guidance Methods}
In practice, vanilla diffusion models produce plausible samples, but with no regard to specific user requirements or objectives. Thus, guidance in diffusion models is commonly introduced to steer the generative process towards specific desired outcomes during the denoising process \cite{dhariwal2021diffusion}. In this approach, a separate model (e.g., neural network) which is able to predict the objective value of the condition is assumed to be available, such that its gradients can be used to steer the denoising process towards a target criteria. Some related research can be found in \cite{wallace2023end,kim2022guided}.




For a diffusion model with an unconditional denoising process $p_{\boldsymbol{\theta}}(\boldsymbol{x}_{t} | \boldsymbol{x}_{t+1})$, \cite{ho2020denoising} shows that, under reasonable assumptions, we can model the distribution as a Gaussian $p_{\boldsymbol{\theta}}(\boldsymbol{x}_{t} | \boldsymbol{x}_{t+1})=\mathcal{N}(\boldsymbol{x}_{t};\boldsymbol{\mu}_{\boldsymbol{\theta}}(\boldsymbol{x}_{t+1}),\boldsymbol{\Sigma}_{\boldsymbol{\theta}}(\boldsymbol{x}_{t+1}))$, where the mean $\boldsymbol{\mu}_{\boldsymbol{\theta}}(\boldsymbol{x}_{t+1})$ and covariance $\boldsymbol{\Sigma}_{\boldsymbol{\theta}}(\boldsymbol{x}_{t+1})$ are modeled by two neural networks. Crucially, a guidance process can be incorporated to post hoc bias the denoising process, and \cite{maze2023diffusion} shows that the denoising guidance can in turn be formulated as a gradient-based update to the denoising Gaussian's mean as

\vspace{0mm}
\begin{equation}
    \mu_{\boldsymbol{\theta}}^c(\boldsymbol{x}_{t}) = \mu_{\boldsymbol{\theta}}(\boldsymbol{x}_{t}) + \alpha \boldsymbol{\Sigma}_{\boldsymbol{\theta}}(\boldsymbol{x}_t) \nabla_{\boldsymbol{x}_t} f(\boldsymbol{x}_t)
\label{conditional update}
\end{equation}

\noindent where $f$ is a differentiable guidance function (e.g., a regressor) that evaluates the denoising outcome; and $\alpha$ is the gradient scaling factor.

Note that the gradient-based guidance mechanisms typically require a differentiable proxy. However, models with black-box characteristics, e.g., the finite element analysis (FEA) software Abaqus and computational fluid dynamics (CFD) software Ansys Fluent commonly used for structural and fluid analysis respectively, can evaluate objective functions but are inherently non-differentiable. This lack of differentiability makes them incompatible with gradient-based guidance mechanisms and limits their integration into classifier/regressor-guided diffusion frameworks. To address this limitation, we propose an evolvable conditional diffusion method to eliminate the need for differentiable proxies.

\vspace{-0mm}
\section{Gradient-Free Conditional Diffusion Algorithm}
\subsection{Theoretical Background}
In this section, we derive a gradient-free guidance approach from first principles through the lens of probabilistic, population-based evolutionary computation. We formulate the guidance as an optimization problem where the denoising process is evolved to advance in the direction that maximizes the fitness function $f$, thereby updating the parameters of the denoising distribution. In probabilistic evolution, the optimization problem can be reformulated as maximizing the following expected fitness under the underlying population distribution model \cite{ollivier2017information,gupta2022half}

\vspace{0mm}
\begin{equation}
    J(\boldsymbol{\omega}) = \mathbb{E}_{\boldsymbol{\omega}}[f(\boldsymbol{x}_t)] = \int f(\boldsymbol{x}_t) p(\boldsymbol{x}_t | \boldsymbol{\omega})  d\boldsymbol{x}_t
\label{optimization objective}
\end{equation}

\noindent where $f$ is the black-box guidance function; $\boldsymbol{\omega} = (\boldsymbol{\mu_{\boldsymbol{\theta}}}, \boldsymbol{\Sigma_{\boldsymbol{\theta}}}$) represents the denoising distribution parameters, which are evolved to produce samples that are more desirable as assessed by the fitness function; $\boldsymbol{x}_t$ represents individual samples drawn from the denoising distribution. In our proposed gradient-free approach where we maximize $J(\boldsymbol{\omega})$, $f$ no longer has the same requirements for differentiability as the equivalent $f$ in Eq. (\ref{conditional update}).

\begin{proposition}

Given the optimization objective $J(\boldsymbol{\omega}) = \mathbb{E}_{\boldsymbol{\omega}}[f(\boldsymbol{x}_t)]$; $\boldsymbol{\omega} = (\boldsymbol{\mu_{\boldsymbol{\theta}}}, \boldsymbol{\Sigma_{\boldsymbol{\theta}}}$). The gradient-free update rule to the mean of the denoising distribution is given by

\vspace{0mm}
\begin{equation}
    \boldsymbol{\mu}_{\boldsymbol{\theta}}^c = \boldsymbol{\mu_{\theta}} + \alpha \widetilde{\nabla}_{\boldsymbol{\mu_{\theta}}}J(\boldsymbol{\omega})
\label{eqn:gradient-free-update}
\end{equation}
\noindent where $\widetilde{\nabla}_{\boldsymbol{\mu_{\theta}}}J(\boldsymbol{\omega})$ denotes the natural gradient of $J(\boldsymbol{\omega})$ with respect to the mean $\boldsymbol{\mu_{\theta}}$. Then, when $||\boldsymbol{\Sigma_{\boldsymbol{\theta}}}|| \rightarrow 0$, we have $\widetilde{\nabla}_{\boldsymbol{\mu_{\theta}}}J(\boldsymbol{\omega}) = \boldsymbol{\Sigma}_{\boldsymbol{\theta}} \nabla_{\boldsymbol{x}_t} f(\boldsymbol{x}_t)$ which directly corresponds to the gradient-based update in Eq. (\ref{conditional update}).

\end{proposition}

\renewcommand\qedsymbol{$\blacksquare$}

\begin{proof}
The natural gradient to the objective $J(\boldsymbol{\omega})$ can be formalized as the solution to the constrained optimization problem \cite{wierstra2014natural}
\begin{subequations}
    \begin{align}
        & \max_{\delta \boldsymbol{\omega}} J(\boldsymbol{\omega} + \delta \boldsymbol{\omega}) \approx J(\boldsymbol{\omega}) + \delta \boldsymbol{\omega}^\top \nabla_{\boldsymbol{\omega}}J(\boldsymbol{\omega}) \\
        & s.t. \ D(\boldsymbol{\omega} + \delta \boldsymbol{\omega}\ ||\ \boldsymbol{\omega}) = \epsilon
    \end{align}
\end{subequations}
\noindent where $D$ is the Kullback-Leibler divergence between two probability distributions, and $\epsilon$ is a small increment size. For $\delta \boldsymbol{\omega} \rightarrow 0$, the solution to this can be found using a Lagrangian multiplier, yielding the necessary condition
\begin{equation}
    \boldsymbol{F} \delta \boldsymbol{\omega} = \phi \nabla_{\boldsymbol{\omega}}J(\boldsymbol{\omega})
\end{equation}
\noindent where $\boldsymbol{F} = \mathbb{E} [\nabla_{\boldsymbol{\omega}} \log p(\boldsymbol{x}_t | \boldsymbol{\omega}) \nabla_{\boldsymbol{\omega}} \log p(\boldsymbol{x}_t | \boldsymbol{\omega})^\top]$ is the Fisher information matrix, and for some constant $\phi > 0$. The direction of the natural gradient $\widetilde{\nabla}_{\boldsymbol{\omega}}J(\boldsymbol{\omega})$ is thus given by $\delta \boldsymbol{\omega}$
\begin{equation}
    \widetilde{\nabla}_{\boldsymbol{\omega}}J(\boldsymbol{\omega}) = \boldsymbol{F}^{-1} \nabla_{\boldsymbol{\omega}}J(\boldsymbol{\omega}).
\end{equation}

Given further that $p(\boldsymbol{x}_t | \boldsymbol{\omega})$ is a Gaussian distribution, the Fisher information matrix for $\boldsymbol{\mu_{\theta}}$ is given by
\begin{subequations}
    \begin{align}
        \boldsymbol{F}_{\boldsymbol{\mu_{\theta}}} &= \mathbb{E} [\nabla_{\boldsymbol{\mu_{\theta}}} \log p(\boldsymbol{x}_t | \boldsymbol{\omega}) \nabla_{\boldsymbol{\mu_{\theta}}} \log p(\boldsymbol{x}_t | \boldsymbol{\omega})^\top] \\ 
        &= \mathbb{E} [(\boldsymbol{\Sigma}_{\boldsymbol{\theta}}^{-1}(\boldsymbol{x}_t - \boldsymbol{\mu_{\theta}})) (\boldsymbol{\Sigma}_{\boldsymbol{\theta}}^{-1}(\boldsymbol{x}_t - \boldsymbol{\mu_{\theta}}))^\top] \\
        &= \boldsymbol{\Sigma}_{\boldsymbol{\theta}}^{-1} \mathbb{E} [(\boldsymbol{x}_t - \boldsymbol{\mu_{\theta}}) (\boldsymbol{x}_t - \boldsymbol{\mu_{\theta}})^\top] \boldsymbol{\Sigma}_{\boldsymbol{\theta}}^{-1} \\
        &= \boldsymbol{\Sigma}_{\boldsymbol{\theta}}^{-1} \boldsymbol{\Sigma}_{\boldsymbol{\theta}} \boldsymbol{\Sigma}_{\boldsymbol{\theta}}^{-1} \\
        &= \boldsymbol{\Sigma}_{\boldsymbol{\theta}}^{-1}.
    \end{align}
\end{subequations}

Using this relation, the natural gradient to the objective $J(\boldsymbol{\omega})$ can be expressed as
\begin{subequations}
    \begin{align}
    \widetilde{\nabla}_{\boldsymbol{\mu_{\theta}}}J(\boldsymbol{\omega}) &= \boldsymbol{F}_{\boldsymbol{\mu_{\theta}}}^{-1} \nabla_{\boldsymbol{\mu_{\theta}}}J(\boldsymbol{\omega}) \\
        &= \boldsymbol{\Sigma}_{\boldsymbol{\theta}} \nabla_{\boldsymbol{\mu_{\theta}}} \int f(\boldsymbol{x}_t) p(\boldsymbol{x}_t | \boldsymbol{\omega})  d\boldsymbol{x}_t.
    \end{align}
    \label{gradient 12}
\end{subequations}

Note that as $||\boldsymbol{\Sigma_{\boldsymbol{\theta}}}|| \rightarrow 0$, the density $p(\boldsymbol{x}_t | \boldsymbol{\omega})$ approaches a Dirac delta function. Therefore, the translation property of the Dirac delta function allows us to simplify Eq. (\ref{gradient 12}) as

\vspace{0mm}
\begin{subequations}
    \begin{align}
    \widetilde{\nabla}_{\boldsymbol{\mu_{\theta}}}J(\boldsymbol{\omega}) &=  \boldsymbol{\Sigma}_{\boldsymbol{\theta}} \nabla_{\boldsymbol{\mu_{\theta}}} \int f(\boldsymbol{x}_t) p(\boldsymbol{x}_t | \boldsymbol{\omega})  d\boldsymbol{x}_t \\
        &= \boldsymbol{\Sigma}_{\boldsymbol{\theta}} \nabla_{\boldsymbol{x}_t} f(\boldsymbol{x}_t).
    \end{align}
\end{subequations}

Consequently, the derived gradient-free update mathematically corresponds to the gradient-based update used in conventional guided diffusion models, i.e., Eq. (\ref{conditional update}). 

\end{proof}

\vspace{0mm}
\subsection{Implementation Details}
In practice, the gradient-free update term in Eq. (\ref{eqn:gradient-free-update}) is estimated by Monte Carlo sampling of a population of candidate solutions. By the Leibniz integral rule, we express $\widetilde{\nabla}_{\boldsymbol{\mu_{\theta}}}J(\boldsymbol{\omega})$ in Eq. (\ref{gradient 12}) as
\begin{subequations}
\begin{align}
    \widetilde{\nabla}_{\boldsymbol{\mu_{\theta}}}J(\boldsymbol{\omega}) & = \boldsymbol{\Sigma}_{\boldsymbol{\theta}} \ \nabla_{\boldsymbol{\mu_{\theta}}} \int f(\boldsymbol{x}_t) p(\boldsymbol{x}_t | \boldsymbol{\omega})  d\boldsymbol{x}_t \\ 
    & = \boldsymbol{\Sigma}_{\boldsymbol{\theta}} \int f(\boldsymbol{x}_t) \nabla_{\boldsymbol{\mu_{\theta}}} p(\boldsymbol{x}_t | \boldsymbol{\omega})  d\boldsymbol{x}_t.
\end{align}
\label{natural gradient}
\end{subequations}

By applying the log-likelihood trick, i.e., $\nabla_{\boldsymbol{\mu_{\theta}}} p(\boldsymbol{x}_t | \boldsymbol{\omega}) = \nabla_{\boldsymbol{\mu_{\theta}}} \log p(\boldsymbol{x}_t | \boldsymbol{\omega}) p(\boldsymbol{x}_t | \boldsymbol{\omega})$, Eq. (\ref{natural gradient}) is reformulated as
\begin{subequations}
\begin{align}
    \widetilde{\nabla}_{\boldsymbol{\mu_{\theta}}}J(\boldsymbol{\omega}) & = \boldsymbol{\Sigma}_{\boldsymbol{\theta}}  \int f(\boldsymbol{x}_t) \nabla_{\boldsymbol{\mu_{\theta}}} \log p(\boldsymbol{x}_t | \boldsymbol{\omega}) p(\boldsymbol{x}_t | \boldsymbol{\omega}) d\boldsymbol{x}_t \label{subeqn:loglik} \\
    & \approx \boldsymbol{\Sigma}_{\boldsymbol{\theta}} \frac{1}{N_s} \sum_{i = 1}^{N_s} f(\boldsymbol{x}_t^i) \nabla_{\boldsymbol{\mu_{\theta}}} \log p(\boldsymbol{x}_t^i | \boldsymbol{\omega}) \label{subeqn:montecarlo} \\
    &= \boldsymbol{\Sigma}_{\boldsymbol{\theta}} \frac{1}{N_s} \sum_{i = 1}^{N_s} f(\boldsymbol{x}_t^i) \boldsymbol{\Sigma}_{\boldsymbol{\theta}}^{-1} (\boldsymbol{x}_t^i - \boldsymbol{\mu_{\theta}}) \label{subeqn:gradexpression} \\
    &= \frac{1}{N_s} \sum_{i = 1}^{N_s} f(\boldsymbol{x}_t^i)  (\boldsymbol{x}_t^i - \boldsymbol{\mu_{\theta}}). \label{subeqn:naturalgradient}
\end{align}
\label{natural gradient1}
\end{subequations}

Note that Eq. (\ref{subeqn:montecarlo}) is approximated by the Monte Carlo estimate of the search gradient, which is obtained without ever having to compute the derivatives of the guidance signal $f$. 


Since Eq. (\ref{subeqn:naturalgradient}) is sensitive to the magnitude and extreme values of the fitness function, a fitness shaping \cite{wierstra2014natural} approach is applied to transform the actual fitness values into rank values. This replaces $f$ with a rank-based function $r$ as 
\vspace{0mm}
\begin{equation}
    r(\boldsymbol{x}_t) = a + b \cdot {\rm rank}(f(\boldsymbol{x}_t))
\label{fitness shaping}
\end{equation}

\noindent where ${\rm rank}(\boldsymbol{x}_t)$ denotes the rank values of all samples in $\boldsymbol{x}_t$, determined based on the actual fitness value $f(\boldsymbol{x}_t)$, and a lower rank value indicates a smaller actual fitness value; $a$ and $b$ are constants used for rescaling. It is evident that Eq. (\ref{fitness shaping}) controls the range of fitness values and thereby ensures the invariance of our method to order-preserving fitness transformations. As the fitness function $f$ is now replaced by a rank-based function $r$, the natural gradient in Eq. (\ref{subeqn:naturalgradient}) can be further expressed as
\begin{subequations}
    \begin{align}
         \widetilde{\nabla}_{\boldsymbol{\mu_{\theta}}}J(\boldsymbol{\omega}) 
        &\approx \frac{1}{N_s} \sum_{i = 1}^{N_s} r(\boldsymbol{x}_t^i)  (\boldsymbol{x}_t^i - \boldsymbol{\mu_{\theta}}) \\
        & = \frac{1}{N_s} \boldsymbol{M}_t \boldsymbol{R}_t
        \label{updates for gd}
    \end{align}
\end{subequations}

\noindent where $\boldsymbol{M}_t = [m_{ni}]_{N \times N_s}$ with $m_{ni} := x_{tn}^i - \mu_{\boldsymbol{\theta}}^n$, where $N$ is the dimensionality of the samples, and $N_s$ is the number of samples drawn from the current distribution $(\boldsymbol{\mu}_{\boldsymbol{\theta}}(\boldsymbol{x}_t), \; \boldsymbol{\Sigma}_{\boldsymbol{\theta}}(\boldsymbol{x}_t))$; $\boldsymbol{R}_t = [r_t^1, \cdots, r_t^{N_s}]^{\top}$. 

\vspace{0mm}
\subsection{Algorithm}
\vspace{0mm}

The steps of the evolvable conditional diffusion method are outlined in Algorithm \ref{alg:algorithm}, assuming that a pre-trained diffusion model parameterized by $\boldsymbol{\theta}$ that predicts $(\boldsymbol{\mu}_{\boldsymbol{\theta}}, \; \boldsymbol{\Sigma}_{\boldsymbol{\theta}})$ is available. 

\begin{algorithm}[tb]
    \caption{Evolvable conditional diffusion method}
    \label{alg:algorithm}
    \textbf{Input}: number of denoising step $T$, number of samples for gradient estimation $N_s$, dimensionality of the noised sample $N$, gradient scaling factor $\alpha$, unconditional diffusion model $(\boldsymbol{\mu}_{\boldsymbol{\theta}}, \; \boldsymbol{\Sigma}_{\boldsymbol{\theta}})$, and model for evaluating the fitness $f(\cdot)$\\
    \textbf{Output}: final guided sample $\boldsymbol{x}_0$
    \begin{algorithmic}[1] 
        \STATE $\boldsymbol{x}_T$ $\leftarrow$ sampling from $\mathcal{N}(0, \boldsymbol{I})$
        \FOR{$t \in [T, T-1, \cdots, 1]$}
            \FOR{$i \in [1, 2, \cdots, N_s]$}
            \STATE $\boldsymbol{x}_t^i$ $\leftarrow$ sampling from $\mathcal{N}(\boldsymbol{\mu}_{\boldsymbol{\theta}}(\boldsymbol{x}_t), \; \boldsymbol{\Sigma}_{\boldsymbol{\theta}}(\boldsymbol{x}_t))$
            \STATE $f^i_t \leftarrow f(\boldsymbol{x}_t^i)$
            \ENDFOR
            \STATE $r_t^1, r_t^2, \ldots, r_t^{N_s} \leftarrow \text{FitnessShaping}(f_t^1, f_t^2, \ldots, f_t^{N_s})$
            \STATE $\boldsymbol{M}_t \leftarrow [m_{ni}]_{N \times N_s}$ with $m_{ni} := x_{tn}^i - \mu_{\boldsymbol{\theta}}^n$
            \STATE $\boldsymbol{R}_t \leftarrow [r_t^1, \cdots, r_t^{N_s}]^{\top}$
            \STATE $\boldsymbol{\mu}_{\boldsymbol{\theta}}^c(\boldsymbol{x}_t) \leftarrow \boldsymbol{\mu}_{\boldsymbol{\theta}}(\boldsymbol{x}_t) + \alpha \boldsymbol{M}_t \boldsymbol{R}_t$
            \STATE $\boldsymbol{x}_{t-1} \leftarrow$ sampling from $\mathcal{N}(\boldsymbol{\mu}^c_{\boldsymbol{\theta}}(\boldsymbol{x}_t), \boldsymbol{\Sigma}_{\boldsymbol{\theta}}(\boldsymbol{x}_t))$
        \ENDFOR
        \STATE \textbf{return} $\boldsymbol{x}_0$
    \end{algorithmic}
\end{algorithm}

The initial distribution for the evolutionary optimization problem is given by $\boldsymbol{\omega} = \boldsymbol{\omega}_t = (\boldsymbol{\mu}_{\boldsymbol{\theta}}(\boldsymbol{x}_t), \; \boldsymbol{\Sigma}_{\boldsymbol{\theta}}(\boldsymbol{x}_t))$, and $N_s$ samples are drawn from the distribution $\mathcal{N}(\boldsymbol{\mu}_{\boldsymbol{\theta}}(\boldsymbol{x}_t), \; \boldsymbol{\Sigma}_{\boldsymbol{\theta}}(\boldsymbol{x}_t))$ in Line 4. The actual fitness values for these samples are computed accordingly in Line 5. To decrease the influence of extreme values, rank-based fitness shaping as per Eq. (\ref{fitness shaping}) is applied to transform the actual fitness values to rank values. A single gradient ascent step at $\boldsymbol{\omega}_t$ based on Eq. (\ref{updates for gd}) is then performed with a gradient scaling factor $\alpha$ to update the denoising distribution in the direction of higher fitness values in Line 10. The conditional denoising mean $\boldsymbol{\mu}_{\boldsymbol{\theta}}^c(\boldsymbol{x}_t)$ is now obtained. With conditional sampling, a new sample for the next denoising step is generated in Line 11. Subsequently, the distribution parameters for the next denoising step $\boldsymbol{\omega}_{t-1} = (\boldsymbol{\mu}_{\boldsymbol{\theta}}(\boldsymbol{x}_{t-1}), \; \boldsymbol{\Sigma}_{\boldsymbol{\theta}}(\boldsymbol{x}_{t-1}))$ are determined, and the initial $\boldsymbol{\omega}$ for this step is then set using $\boldsymbol{\omega}_{t-1}$. This process is repeated iteratively until the final denoising state $\boldsymbol{x}_0$ is reached.


As the updates are approximated directly through the use of drawn samples from the evolved distribution and corresponding rank values, this method is applicable even when the gradients of the fitness function evaluator are difficult to obtain (e.g., in the use of non-differentiable, black-box numerical solvers). 


\vspace{-0mm}
\section{Experiments}
\subsection{Design of Fluidic Channel Topology}
\textbf{Problem Statement.} In this section, we validate the effectiveness of our proposed evolvable conditional diffusion on the design of fluidic channel topology, a fundamental problem with implications on the performance of heat exchangers across multiple domains, including semiconductor chip cooling, battery pack liquid cooling, and chemical reactor cooling. Specifically, we seek to optimize the geometry of fluid channels to minimize pressure drop across the inlet and outlet ($\Delta p$) as pressure drop is a proxy for the amount of energy required to drive flow through the system, and designs with reduced pressure drop typically have superior performance in terms of energy efficiency. 

This pressure drop can be obtained conventionally by solving the steady-state incompressible Navier-Stokes (N-S) equations as per Eq. (\ref{Eq:2D_NS}) \cite{wei2023select}

\vspace{0mm}
\begin{equation}
\label{Eq:2D_NS}
    \nabla \cdot \vec{u} = 0  \quad \quad \quad \\
    (\vec{u} \cdot \nabla)\vec{u} = \frac{1}{\rm Re} \nabla^2 \vec{u} -\nabla p
\end{equation}



\noindent where $\vec{u}$ represents the velocity vector; $p$ denotes the pressure; ${\rm Re} = 500$ is the Reynolds number under investigation.

\begin{figure*}[t]
\centering
\vspace*{0mm}
\includegraphics[width=1\textwidth]{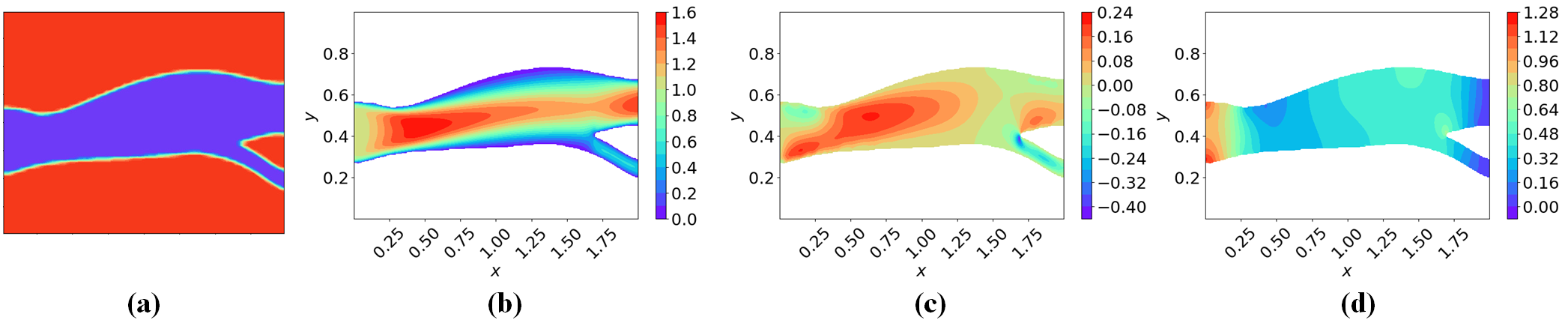}
\vspace*{-0mm}
\caption{CFD results for a representative topology. (a) Channel topology (i.e., the final denoising state $\boldsymbol{x}_0$ generated by the diffusion model). (b) $u$-velocity contour obtained by CFD. (c) $v$-velocity contour obtained by CFD. (d) $p$ contour obtained by CFD.}
\vspace*{-0mm}
\label{Fig_cfd}
\end{figure*}

\noindent \textbf{Implementation Details.} For demonstration of the evolvable conditional diffusion model, we first assume availability of a black-box solver that can provide evaluations of the pressure drop when provided a specific channel topology and a pre-trained diffusion model that can generate random fluidic channel topologies. Examples of such black-box solvers include CFD numerical solvers such as Ansys Fluent. During the denoising process, evaluations of the objective (i.e., $\Delta p$) corresponding to the final denoising state $\boldsymbol{x}_0$ can be obtained by calling this black-box solver. Figure \ref{Fig_cfd} presents sample CFD results for a representative fluidic topology. As a proof-of-concept, a regressor was trained to provide the black-box fitness evaluation using a paired dataset comprising topology designs and their corresponding $\Delta p$. Using the pre-trained diffusion model, guidance is applied for the second half of the denoising process with 30 samples evaluated per denoising step for gradient estimation. The entire denoising process consists of 100 steps, and the input design representation has a spatial resolution of $64 \times 64$.

\begin{figure}[t]
\centering
\vspace*{0mm}
\includegraphics[width=0.48\textwidth]{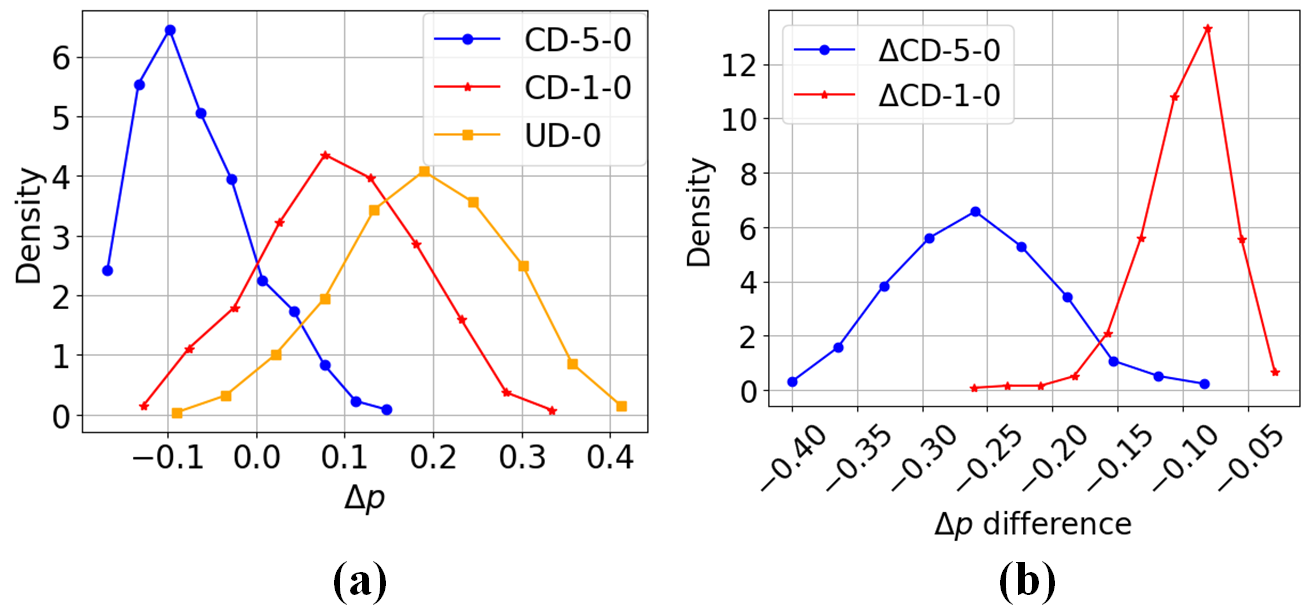}
\vspace*{-0mm}
\caption{Histograms of $\Delta p$ and the difference in $\Delta p$ between samples generated with 50-step guidance and without guidance for 1000 test samples in fluidic channel topology design. During the denoising process, $\Delta p$ is normalized by $\ln{\Delta p}/5$. (a) Histogram of $\Delta p$. ``CD-5-0'' represents $\Delta p$ for samples generated with guidance and $\alpha = 5$. ``CD-1-0'' represents $\Delta p$ for samples generated with guidance and $\alpha = 1$. ``UD-0'' represents $\Delta p$ for samples generated without guidance. (b) Histogram of the difference in $\Delta p$ across paired samples generated with and without guidance. ``$\Delta$CD-5-0'' represents the difference in $\Delta p$ between ``CD-5-0'' and ``UD-0''. ``$\Delta$CD-1-0'' represents the difference in $\Delta p$ between ``CD-1-0'' and ``UD-0''.}
\vspace*{-0mm}
\label{Fig_deltap_histogram}
\end{figure}

\noindent \textbf{Results.} We generate 1000 samples for assessment of the effectiveness of our proposed method. The histogram of $\Delta p$ obtained with and without gradient-free guidance during denoising across these 1000 samples is presented in Figure \ref{Fig_deltap_histogram} (a). The results indicate that the distribution of $\Delta p$ of the samples generated with guidance is significantly lower compared to the baseline generated samples. Figure \ref{Fig_deltap_histogram} (b) further illustrates that all 1000 samples obtained via conditional diffusion have reduced $\Delta p$ relative to the corresponding sample generated without guidance in the denoising process. Furthermore, increasing the gradient scaling factor $\alpha$ clearly biases the denoising process towards designs with even lower $\Delta p$, demonstrating the potential for generating designs which better satisfy one's criteria. The results highlight the effectiveness of the proposed evolvable conditional diffusion in scenarios where the solver is non-differentiable.

\begin{figure}[t]
\centering
\vspace*{-0mm}
\includegraphics[width=0.48\textwidth]{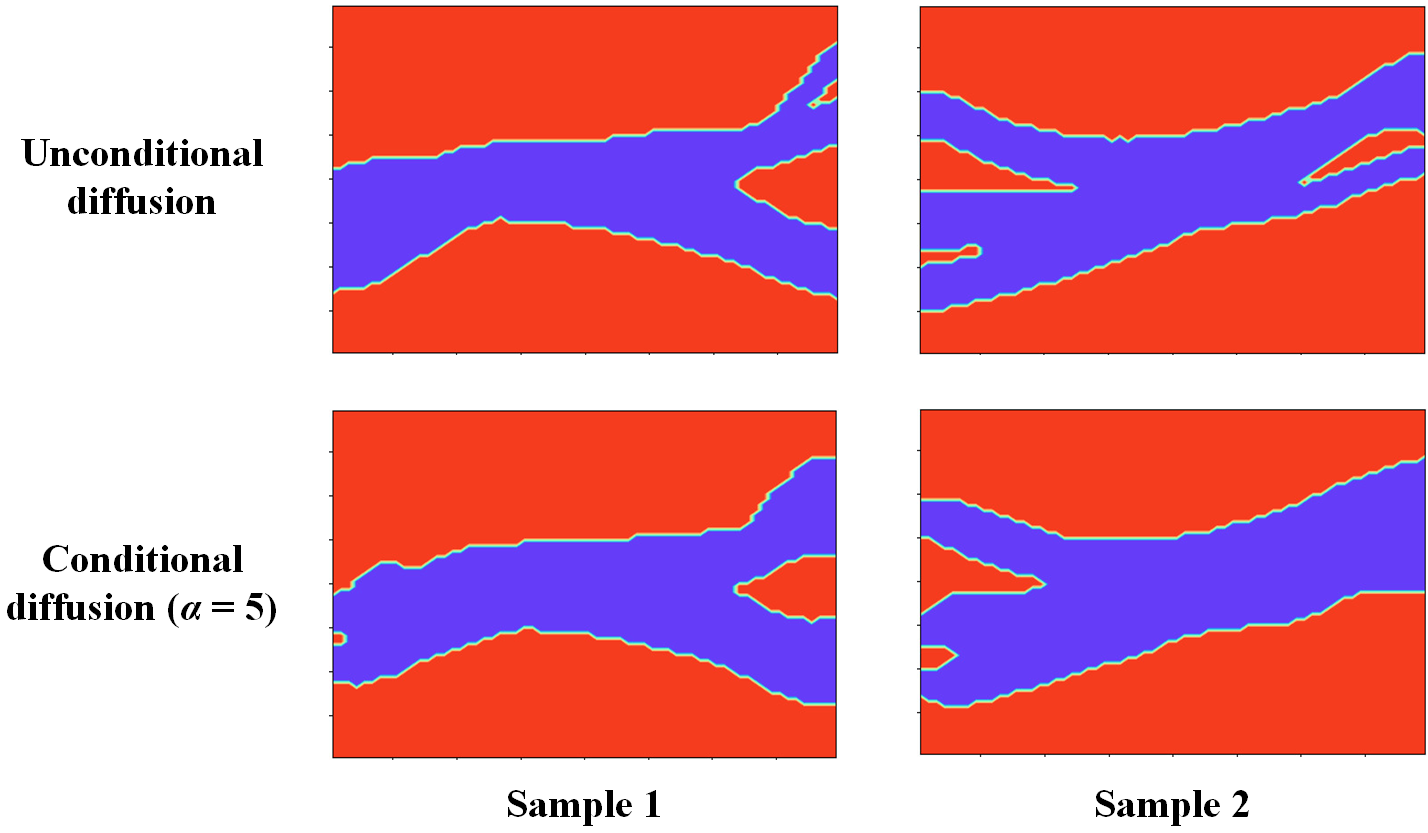}
\vspace*{-0mm}
\caption{Examples of fluidic channel topologies generated with and without guidance.}
\vspace*{-0mm}
\label{Fig_x0}
\end{figure}

\begin{figure}[t]
\centering
\vspace*{0mm}
\includegraphics[width=0.48\textwidth]{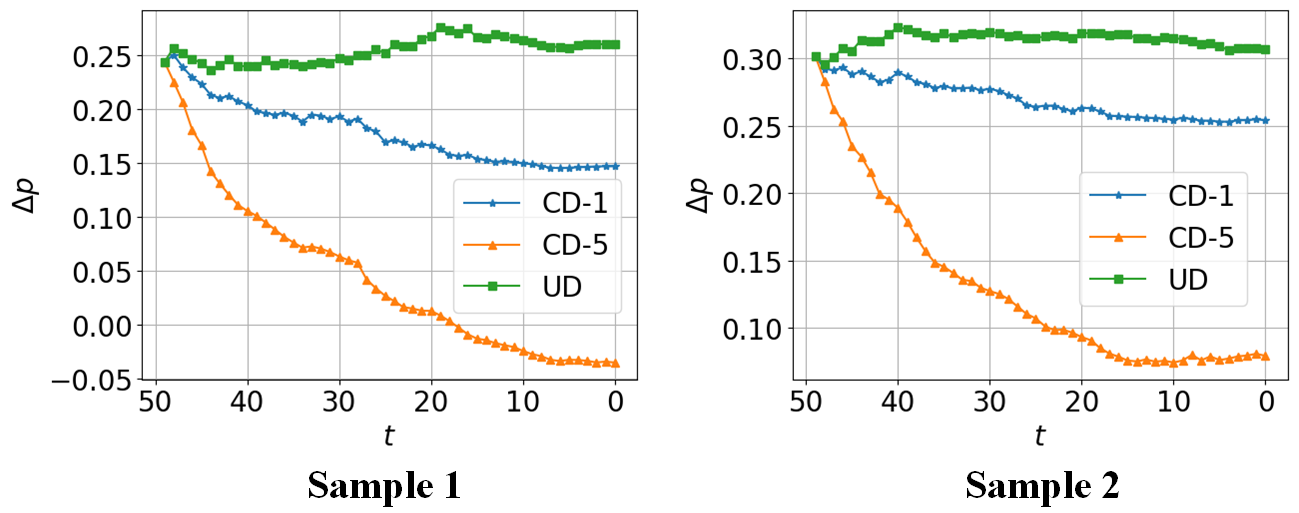}
\vspace*{0mm}
\caption{Curves illustrating the change in $\Delta p$ for the final 50 denoising steps of generation with and without guidance. ``CD-5'', ``CD-1'' and  ``UD'' indicate $\Delta p$ is plotted for the scenarios where guidance with a scaling factor of $\alpha = 5$, $\alpha = 1$, and $\alpha = 0$ respectively, are applied.}
\vspace*{0mm}
\label{Fig_p_curve}
\end{figure}

Two representative samples from the 1000 generated samples are presented in Figure \ref{Fig_x0} to illustrate the final images (i.e., $\boldsymbol{x}_0$) generated with and without guidance. Notably, when $\alpha = 5$, the outlet area is observed to be widened, causing a reduction in $\Delta p$ compared to the baseline design. $\Delta p$ curves for the final 50 denoising steps of conditional and unconditional diffusion are presented in Figure \ref{Fig_p_curve}. When the guidance is applied during denoising, $\Delta p$ decreases as denoising progresses, with a more significant drop observed for larger $\alpha$. In contrast, $\Delta p$ in the baseline denoising process remains fairly consistent, further illustrating the effectiveness of the proposed evolvable conditional diffusion.

\begin{figure}[t]
\centering
\vspace*{0mm}
\includegraphics[width=0.48\textwidth]{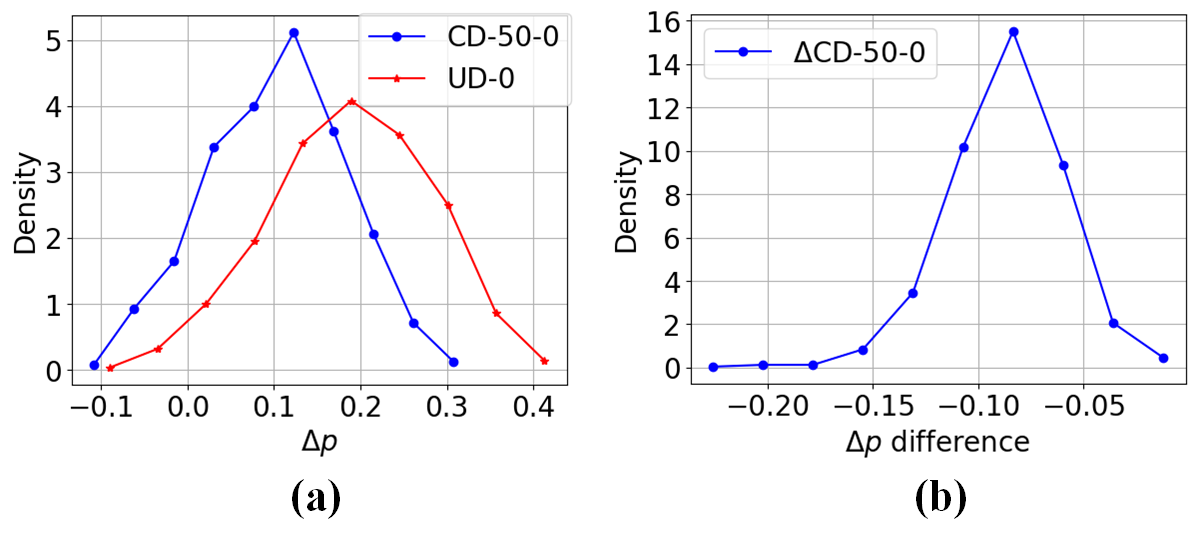}
\vspace*{0mm}
\caption{Histograms of $\Delta p$ and the difference in $\Delta p$ between samples generated with 10-step guidance and without guidance for 1000 test samples in fluidic channel topology design. During the denoising process, $\Delta p$ is normalized by $\ln{\Delta p}/5$. (a) Histogram of $\Delta p$. ``CD-50-0'' represents $\Delta p$ for samples generated with $\alpha = 50$. ``UD-0'' represents $\Delta p$ for samples generated without guidance. (b) Histogram of the difference in $\Delta p$ across the paired samples generated with and without guidance. ``$\Delta$CD-50-0'' represents the difference in $\Delta p$ between ``CD-50-0'' and ``UD-0''.}
\vspace*{0mm}
\label{Fig_deltap_histogram1}
\end{figure}

To further demonstrate the effectiveness of the proposed approach, we applied guidance for a smaller fraction of the denoising steps in the generation process. Specifically, the gradient-free guidance was applied only for the last 10 steps (as opposed to the previously presented 50 steps). The histogram of $\Delta p$ and difference in $\Delta p$ between samples generated with and without guidance are presented for 1000 generated samples in Figure \ref{Fig_deltap_histogram1}. The results show that $\Delta p$ for the 1000 samples generated with guidance applied during denoising for just 10 steps remain significantly lower than the baseline, further demonstrating the method's effectiveness.

\vspace{0mm}
\subsection{Meta-Surface Design}
\begin{figure}[t]
\centering
\vspace*{0mm}
\includegraphics[width=0.48\textwidth]{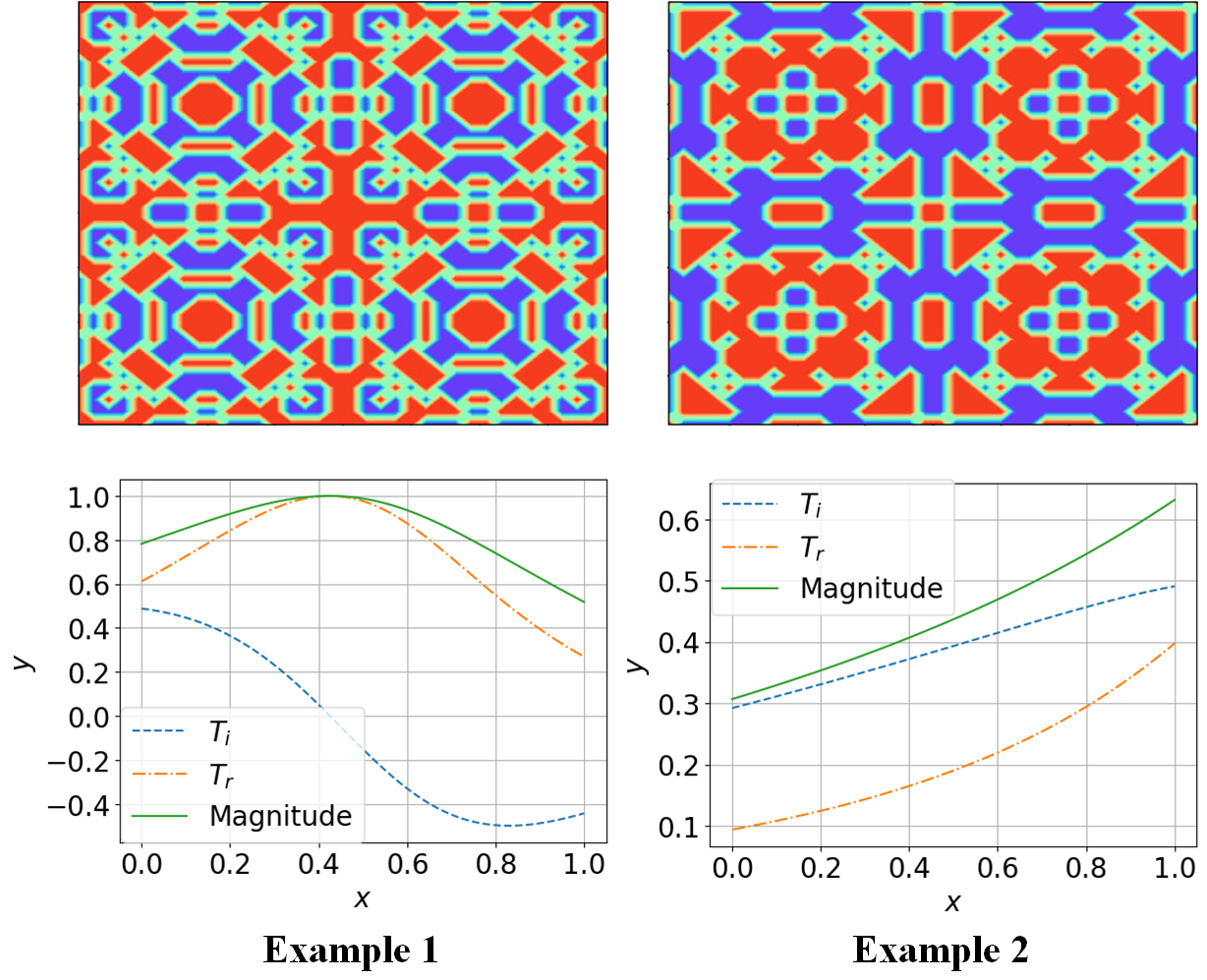}
\vspace*{0mm}
\caption{Two representative meta-surface designs along with curves depicting the real and imaginary components of their transmission profiles (i.e., $T_r$ and $T_i$ respectively), and the corresponding transmission magnitudes.}
\vspace*{0mm}
\label{Fig_examples}
\end{figure}

\textbf{Problem Statement.} In this section, we demonstrate the effectiveness of the proposed guidance for design of frequency-selective meta-surfaces. Meta-surface design plays a crucial role in modern electronic devices, including 5G telecommunications. Electrical engineers typically rely on high-fidelity electromagnetic simulation tools (e.g., Ansys HFSS) in order to obtain and analyze electromagnetic behavior of these complex surfaces \cite{yang2022physics}. These simulators solve Maxwell's equations to provide the transmission and reflectance responses of various meta-surface designs. As an example, a meta-surface can be designed to match a given profile of the real ($T_r$) and imaginary ($T_i$) components of the transmission. A sample band-pass-type profile, where the magnitude of $T_r$ and $T_i$ must resemble the parabolic profile $y = 1 - 2 \times (x - 0.5)^2$ across a specific frequency range, is one of many possible common design specifications engineers encounter. In this instance, the mean absolute error (MAE) between the predicted and target magnitudes of $T_r$ and $T_i$ is a key design goal, and minimizing MAE can be a key metric for guiding the design process. We present two representative meta-surface designs along with their corresponding $T_r$, $T_i$, and magnitude of $T_r$ and $T_i$ curves in Figure \ref{Fig_examples}.

\noindent \textbf{Implementation Details.} Following the previous example, we assume the availability of a black-box solver capable of evaluating the transmission profile for a given meta-surface design, as well as a pre-trained diffusion model that can generate random meta-surface designs. Examples of such black-box solvers include typical computational electromagnetics numerical solvers such as Ansys HFSS. During the denoising process, evaluations of the objective (i.e., MAE relative to the desired target profile) corresponding to the final design $\boldsymbol{x}_0$ can be obtained by calling this black-box solver. As a proof-of-concept, a regressor is trained to provide the black-box fitness evaluation using a paired dataset comprising meta-surface designs and their corresponding transmission profiles. 
Using the pre-trained diffusion model, guidance is applied for the second half of the denoising process with 30 samples evaluated per denoising step for gradient estimation. The entire denoising process consists of 100 steps, and the input design representation has a spatial resolution of $64 \times 64$.

\begin{figure}[t]
\centering
\vspace*{0mm}
\includegraphics[width=0.48\textwidth]{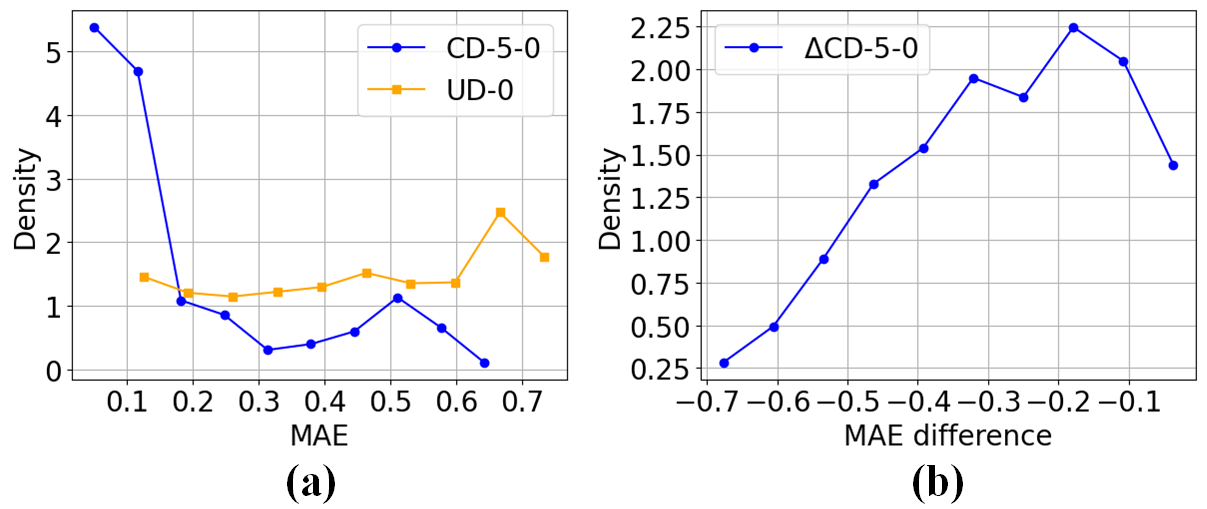}
\vspace*{0mm}
\caption{Histograms of MAE and the difference in MAE between samples generated with 50-step guidance and without guidance for 1000 test samples in meta-surface design. (a) Histogram of MAE. ``CD-5-0'' represents MAE of meta-surface designs generated using $\alpha = 5$. ``UD-0'' represents MAE of meta-surface designs generated without any guidance during the denoising process. (b) Histogram of the difference in MAE across the paired samples generated with and without guidance. ``$\Delta$CD-5-0'' represents the difference in MAE between ``CD-5-0'' and ``UD-0''.}
\vspace*{0mm}
\label{Fig_MAE_histogram}
\end{figure}

\noindent \textbf{Results.} We generate 1000 samples to evaluate the effectiveness of the proposed evolvable conditional diffusion method. The histogram of the MAE and the difference in MAE between samples generated with and without guidance is presented in Figure \ref{Fig_MAE_histogram}. Figure \ref{Fig_MAE_histogram} (a) clearly shows that the samples generated with guidance during denoising better match the target profile (i.e., show lower MAE values) than samples generated without guidance. The median MAE of samples generated with guidance is $\approx 0.1$, indicating a good match to the target profile. Figure \ref{Fig_MAE_histogram} (b) further shows that the MAE of meta-surface designs generated with guidance is consistently lower than the baseline samples across all 1000 generated samples, emphasizing the effectiveness of the proposed gradient-free guidance.

\begin{figure}[t]
\centering
\vspace*{0mm}
\includegraphics[width=0.48\textwidth]{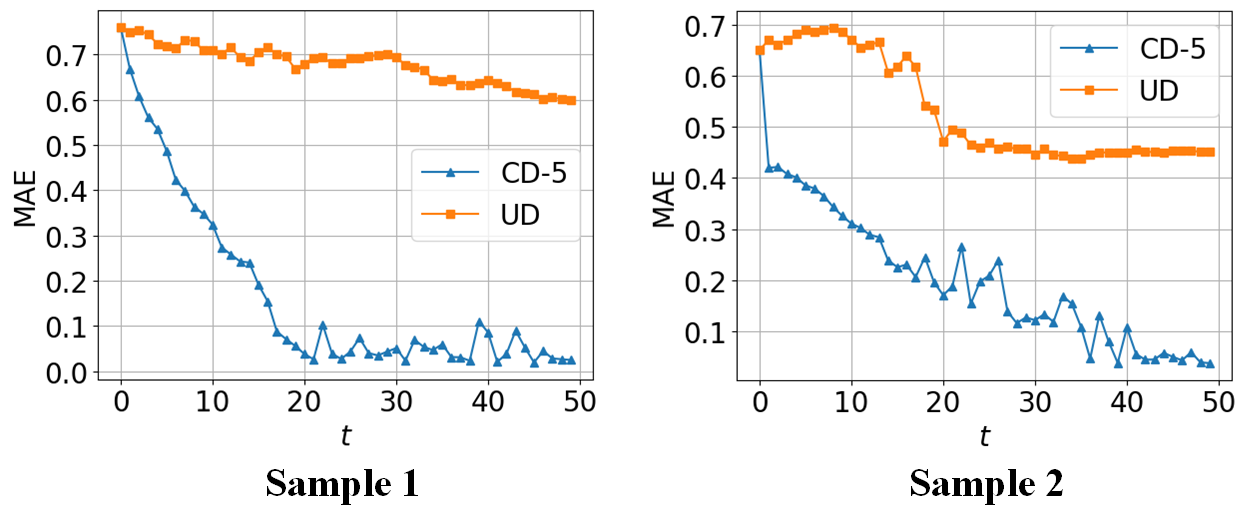}
\vspace*{0mm}
\caption{Curves illustrating the change in MAE for the final 50 denoising steps of generation with and without guidance. ``CD-5'' and  ``UD'' indicate MAE is plotted for the scenarios where guidance with a scaling factor of $\alpha = 5$ and $\alpha = 0$ respectively are applied.}
\vspace*{0mm}
\label{Fig_MAE_curve}
\end{figure}

\begin{figure}[t]
\centering
\vspace*{0mm}
\includegraphics[width=0.48\textwidth]{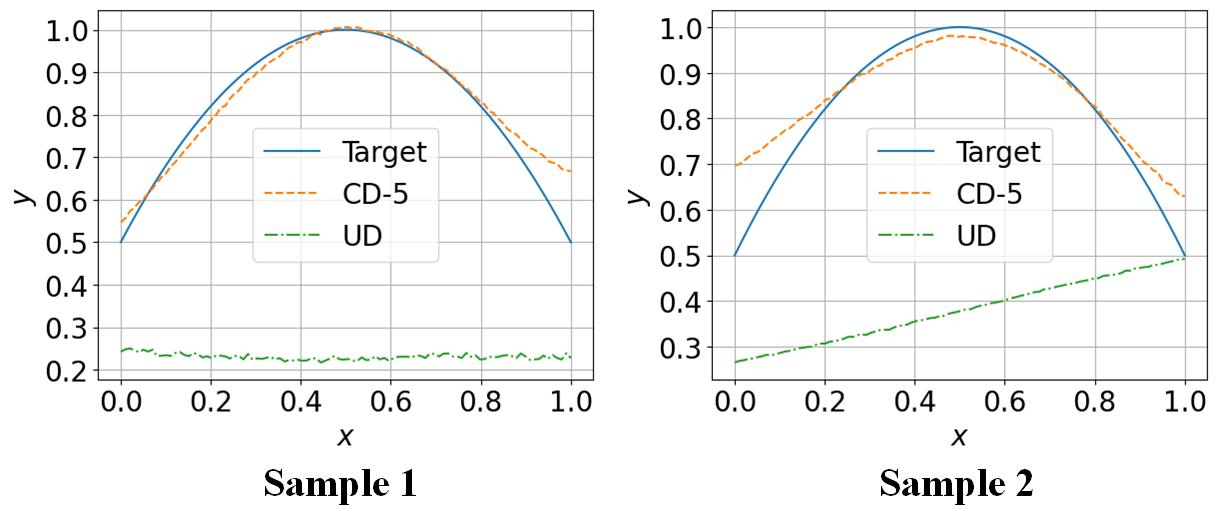}
\vspace*{0mm}
\caption{Predicted transmission profiles of designs generated with and without guidance in the denoising process. ``Target'' indicates target profile which was used as guidance during denoising. ``CD-5'' indicates the predicted profile is obtained through denoising with a guidance of $\alpha = 5$. ``UD'' indicates the predicted profile is obtained with no guidance during denoising.}
\vspace*{0mm}
\label{Fig_target}
\end{figure}

Two samples from the 1000 test samples are selected for detailed analysis. Figure \ref{Fig_MAE_curve} illustrates that the MAE between the transmission profile for samples generated with guidance during denoising and the target profile steadily decreases to a small value, whereas the MAE remains large in the baseline case. Similarly, in Figure \ref{Fig_target}, the transmission profile of a generated meta-surface design aligns closely with the target profile. In contrast, the transmission profile of the baseline design generated without guidance exhibits significant discrepancy from the target. These results highlight the effectiveness of the proposed evolvable conditional diffusion.


\vspace{0mm}
\section{Discussion}
Guidance-based generation with diffusion models offers the advantage of seamless integration of desired performance or criteria into the diffusion denoising process using pre-trained diffusion models. However, a significant drawback is the reliance on differentiable proxies, a rule that precludes the use of many real-world black-box, non-differentiable simulators. This limitation is particularly critical in scientific domains, where accurate evaluation is best achieved by the use of established, physics-based simulators. While a common paradigm is to train a surrogate model of the guiding objective to provide a differentiable approximation of the objective, this often requires thousands of labeled samples which can be prohibitively expensive, particularly for high-dimensional design spaces (e.g., pixel-level geometry parameterizations or meta-surface unit cell designs). While it is possible to utilize finite difference, such approximations are known to scale poorly for high-dimensional inputs common in physical systems. Alternative scalable gradient-free approaches such as the simultaneous perturbation stochastic approximation (SPSA) proposed in \cite{shen2025chemistryinspired} are also known to suffer from high variance (noise) in high-dimensional spaces. 

To address this challenge, we propose an evolvable conditional diffusion method which is fundamentally different from the current paradigm. By integrating evolution strategies directly into the denoising process, we eliminate the need for any \textit{a priori} surrogate model. Further, this approach is more robust to noise as it derives stochastic gradient estimates from a population of sample evaluations. 

More critically, the scalability of our evolvable conditional diffusion algorithm depends on the ``intrinsic" dimensionality of the problem, which is often low even for high-resolution images where only a small number of pixel values affect the target objectives. Empirically, this conforms with our proposed method's strong performance in high-dimensional settings, as the evolution strategies appear to effectively exploit key informative dimensions of the optimization problem, allowing their performance to scale with the intrinsic complexity or effective dimensionality of the task, rather than with the full dimensionality of the input search space \cite{salimans2017evolution}. While the scalability of such evolution strategies was prominently demonstrated by OpenAI in their seminal work \cite{salimans2017evolution}, our method further extends this observation by explicitly incorporating the covariance matrix into the diffusion denoising process. Note that the ability to scale efficiently to high dimensions is yet another explanation of how our method fundamentally departs from recent surrogate-guided diffusion methods \cite{tan2025fast}, where the amount of data needed to train a differentiable surrogate model becomes intractable with increasing problem dimensionality. 


Lastly, while \cite{shen2025chemistryinspired} derive their gradient-free guidance algorithm by adopting and building upon the update rule in \cite{maze2023diffusion}, Proposition 1 shows that this update rule actually emerges from first principles as a mathematical consequence of the probabilistic evolution method, without requiring any prior assumptions. This positions our method as a more principled and scalable alternative for high-dimensional inverse design and optimization tasks, including but not limited to the two examples of fluidic channel topology and meta-surface design as demonstrated in this work.

\vspace{0mm}
\section{Conclusion}

In this work, we propose an evolvable conditional diffusion method by harnessing the generative capacity of probabilistic evolutionary computation. This approach evolves the diffusion process towards denoising distributions with higher fitness, thereby improving the conformance of generated samples to specified, desirable criteria. Notably, the derived update algorithm is provably analogous to the update in standard gradient-based guided diffusion models, but without the need for computing derivatives of the guidance signal. 


We validate the effectiveness of the proposed evolvable conditional diffusion method through two design applications common in engineering: fluidic channel topology design and frequency-selective meta-surface design. Results demonstrate that the proposed method can effectively generate high-quality designs better aligned with desired performance criteria. Furthermore, the proposed method is highly adaptable and can be easily extended to other domains that require optimization through non-differentiable evaluations, as evidenced by our demonstrations in the two distinct fields of fluid dynamics and electromagnetic. This significantly broadens the applicability of diffusion models for design across a multitude of scientific and engineering domains.

\section*{Acknowledgments}
This research is partially supported by the National Research Foundation, Singapore, and Civil Aviation Authority under the Aviation Transformation Programme - ``ATM-Met Integration of Convective Weather Forecast and Impact Forecast Solutions Supporting Singapore Air Traffic Operations" (Award No. ATP2.0\_ATM-MET\_I2R) and A*STAR under BMRC CSF (Award No. C240314053). This research is partially supported by the National Research Foundation, Singapore and DSO National Laboratories under the AI Singapore Programme - ``Design Beyond What You Know: Material-Informed Differential Generative AI (MIDGAI) for Light-Weight High-Entropy Alloys and Multi-functional Composites (Stage 1a)" (AISG Award No. AISG2-GC-2023-010). This research is also partially supported by the Ramanujan Fellowship from the Science and Engineering Research Board, Government of India (Grant No. RJF/2022/000115).

\vspace{-1mm}


\bibliographystyle{named}
\bibliography{ijcai25}

\begin{thebibliography}{}

\bibitem[\protect\citeauthoryear{Avdeyev \bgroup \em et al.\egroup }{2023}]{pmlr-v202-avdeyev23a}
Pavel Avdeyev, Chenlai Shi, Yuhao Tan, Kseniia Dudnyk, and Jian Zhou.
\newblock {D}irichlet diffusion score model for biological sequence generation.
\newblock In {\em Proceedings of the International Conference on Machine Learning}, pages 1276--1301. PMLR, 2023.

\bibitem[\protect\citeauthoryear{Bali \bgroup \em et al.\egroup }{2020}]{bali2020cognizant}
Kavitesh~Kumar Bali, Abhishek Gupta, Yew-Soon Ong, and Puay~Siew Tan.
\newblock Cognizant multitasking in multiobjective multifactorial evolution: {MO-MFEA-II}.
\newblock {\em IEEE Transactions on Cybernetics}, 51(4):1784--1796, 2020.

\bibitem[\protect\citeauthoryear{Bansal \bgroup \em et al.\egroup }{2023}]{bansal2023universal}
Arpit Bansal, Hong-Min Chu, Avi Schwarzschild, Soumyadip Sengupta, Micah Goldblum, Jonas Geiping, and Tom Goldstein.
\newblock Universal guidance for diffusion models.
\newblock In {\em Proceedings of the IEEE/CVF Conference on Computer Vision and Pattern Recognition}, pages 843--852, 2023.

\bibitem[\protect\citeauthoryear{Chen \bgroup \em et al.\egroup }{2024}]{chen2024overview}
Minshuo Chen, Song Mei, Jianqing Fan, and Mengdi Wang.
\newblock An overview of diffusion models: Applications, guided generation, statistical rates and optimization.
\newblock {\em arXiv preprint arXiv:2404.07771}, 2024.

\bibitem[\protect\citeauthoryear{Croitoru \bgroup \em et al.\egroup }{2023}]{croitoru2023diffusion}
Florinel-Alin Croitoru, Vlad Hondru, Radu~Tudor Ionescu, and Mubarak Shah.
\newblock Diffusion models in vision: A survey.
\newblock {\em IEEE Transactions on Pattern Analysis and Machine Intelligence}, 45(9):10850--10869, 2023.

\bibitem[\protect\citeauthoryear{Dhariwal and Nichol}{2021}]{dhariwal2021diffusion}
Prafulla Dhariwal and Alexander Nichol.
\newblock Diffusion models beat {GANs} on image synthesis.
\newblock {\em Advances in Neural Information Processing Systems}, 34:8780--8794, 2021.

\bibitem[\protect\citeauthoryear{Gainza \bgroup \em et al.\egroup }{2020}]{gainza2020deciphering}
Pablo Gainza, Freyr Sverrisson, Frederico Monti, Emanuele Rodola, Davide Boscaini, Michael~M Bronstein, and Bruno~E Correia.
\newblock Deciphering interaction fingerprints from protein molecular surfaces using geometric deep learning.
\newblock {\em Nature Methods}, 17(2):184--192, 2020.

\bibitem[\protect\citeauthoryear{Ghiringhelli \bgroup \em et al.\egroup }{2015}]{ghiringhelli2015big}
Luca~M Ghiringhelli, Jan Vybiral, Sergey~V Levchenko, Claudia Draxl, and Matthias Scheffler.
\newblock Big data of materials science: Critical role of the descriptor.
\newblock {\em Physical Review Letters}, 114(10):105503, 2015.

\bibitem[\protect\citeauthoryear{Gupta \bgroup \em et al.\egroup }{2022}]{gupta2022half}
Abhishek Gupta, Lei Zhou, Yew-Soon Ong, Zefeng Chen, and Yaqing Hou.
\newblock Half a dozen real-world applications of evolutionary multitasking, and more.
\newblock {\em IEEE Computational Intelligence Magazine}, 17(2):49--66, 2022.

\bibitem[\protect\citeauthoryear{Ho \bgroup \em et al.\egroup }{2020}]{ho2020denoising}
Jonathan Ho, Ajay Jain, and Pieter Abbeel.
\newblock Denoising diffusion probabilistic models.
\newblock {\em Advances in Neural Information Processing Systems}, 33:6840--6851, 2020.

\bibitem[\protect\citeauthoryear{Igashov \bgroup \em et al.\egroup }{2024}]{igashov2024equivariant}
Ilia Igashov, Hannes St{\"a}rk, Cl{\'e}ment Vignac, Arne Schneuing, Victor~Garcia Satorras, Pascal Frossard, Max Welling, Michael Bronstein, and Bruno Correia.
\newblock Equivariant 3d-conditional diffusion model for molecular linker design.
\newblock {\em Nature Machine Intelligence}, 6:417–427, 2024.

\bibitem[\protect\citeauthoryear{Kim \bgroup \em et al.\egroup }{2022}]{kim2022guided}
Heeseung Kim, Sungwon Kim, and Sungroh Yoon.
\newblock Guided-{TTS}: A diffusion model for text-to-speech via classifier guidance.
\newblock In {\em International Conference on Machine Learning}, pages 11119--11133. PMLR, 2022.

\bibitem[\protect\citeauthoryear{Kong \bgroup \em et al.\egroup }{2021}]{kong2021diffwave}
Zhifeng Kong, Wei Ping, Jiaji Huang, Kexin Zhao, and Bryan Catanzaro.
\newblock Diffwave: A versatile diffusion model for audio synthesis.
\newblock In {\em International Conference on Learning Representations}, 2021.

\bibitem[\protect\citeauthoryear{Le \bgroup \em et al.\egroup }{2013}]{le2013evolution}
Minh~Nghia Le, Yew~Soon Ong, Stefan Menzel, Yaochu Jin, and Bernhard Sendhoff.
\newblock Evolution by adapting surrogates.
\newblock {\em Evolutionary Computation}, 21(2):313--340, 2013.

\bibitem[\protect\citeauthoryear{Lehman \bgroup \em et al.\egroup }{2020}]{lehman2020surprising}
Joel Lehman, Jeff Clune, Dusan Misevic, Christoph Adami, Lee Altenberg, Julie Beaulieu, Peter~J Bentley, Samuel Bernard, Guillaume Beslon, David~M Bryson, et~al.
\newblock The surprising creativity of digital evolution: A collection of anecdotes from the evolutionary computation and artificial life research communities.
\newblock {\em Artificial Life}, 26(2):274--306, 2020.

\bibitem[\protect\citeauthoryear{Li \bgroup \em et al.\egroup }{2022}]{li2022diffusion}
Xiang Li, John Thickstun, Ishaan Gulrajani, Percy~S Liang, and Tatsunori~B Hashimoto.
\newblock Diffusion-{LM} improves controllable text generation.
\newblock {\em Advances in Neural Information Processing Systems}, 35:4328--4343, 2022.

\bibitem[\protect\citeauthoryear{Li \bgroup \em et al.\egroup }{2024}]{li2024derivative}
Xiner Li, Yulai Zhao, Chenyu Wang, Gabriele Scalia, Gokcen Eraslan, Surag Nair, Tommaso Biancalani, Shuiwang Ji, Aviv Regev, Sergey Levine, et~al.
\newblock Derivative-free guidance in continuous and discrete diffusion models with soft value-based decoding.
\newblock {\em arXiv preprint arXiv:2408.08252}, 2024.

\bibitem[\protect\citeauthoryear{Liu and Thuerey}{2024}]{liu2024uncertainty}
Qiang Liu and Nils Thuerey.
\newblock Uncertainty-aware surrogate models for airfoil flow simulations with denoising diffusion probabilistic models.
\newblock {\em AIAA Journal}, 62(8):2912--2933, 2024.

\bibitem[\protect\citeauthoryear{Lyu \bgroup \em et al.\egroup }{2024}]{lyu2024covariance}
Yueming Lyu, Kim~Yong Tan, Yew~Soon Ong, and Ivor Tsang.
\newblock Covariance-adaptive sequential black-box optimization for diffusion targeted generation.
\newblock {\em arXiv preprint arXiv:2406.00812}, 2024.

\bibitem[\protect\citeauthoryear{Maz{\'e} and Ahmed}{2023}]{maze2023diffusion}
Fran{\c{c}}ois Maz{\'e} and Faez Ahmed.
\newblock Diffusion models beat {GANs} on topology optimization.
\newblock In {\em Proceedings of the AAAI Conference on Artificial Intelligence}, volume~37, pages 9108--9116, 2023.

\bibitem[\protect\citeauthoryear{Miikkulainen and Forrest}{2021}]{miikkulainen2021biological}
Risto Miikkulainen and Stephanie Forrest.
\newblock A biological perspective on evolutionary computation.
\newblock {\em Nature Machine Intelligence}, 3(1):9--15, 2021.

\bibitem[\protect\citeauthoryear{Ollivier \bgroup \em et al.\egroup }{2017}]{ollivier2017information}
Yann Ollivier, Ludovic Arnold, Anne Auger, and Nikolaus Hansen.
\newblock Information-geometric optimization algorithms: A unifying picture via invariance principles.
\newblock {\em Journal of Machine Learning Research}, 18(18):1--65, 2017.

\bibitem[\protect\citeauthoryear{Rombach \bgroup \em et al.\egroup }{2022}]{rombach2022high}
Robin Rombach, Andreas Blattmann, Dominik Lorenz, Patrick Esser, and Bj{\"o}rn Ommer.
\newblock High-resolution image synthesis with latent diffusion models.
\newblock In {\em Proceedings of the IEEE/CVF Conference on Computer Vision and Pattern Recognition}, pages 10684--10695, 2022.

\bibitem[\protect\citeauthoryear{Salimans \bgroup \em et al.\egroup }{2017}]{salimans2017evolution}
Tim Salimans, Jonathan Ho, Xi~Chen, Szymon Sidor, and Ilya Sutskever.
\newblock Evolution strategies as a scalable alternative to reinforcement learning.
\newblock {\em arXiv preprint arXiv:1703.03864}, 2017.

\bibitem[\protect\citeauthoryear{Shen \bgroup \em et al.\egroup }{2025}]{shen2025chemistryinspired}
Yuchen Shen, Chenhao Zhang, Sijie Fu, Chenghui Zhou, Newell Washburn, and Barnabas Poczos.
\newblock Chemistry-inspired diffusion with non-differentiable guidance.
\newblock In {\em The Thirteenth International Conference on Learning Representations}, 2025.

\bibitem[\protect\citeauthoryear{Song and Ermon}{2019}]{song2019generative}
Yang Song and Stefano Ermon.
\newblock Generative modeling by estimating gradients of the data distribution.
\newblock {\em Advances in Neural Information Processing Systems}, 32, 2019.

\bibitem[\protect\citeauthoryear{Song and Ermon}{2020}]{song2020improved}
Yang Song and Stefano Ermon.
\newblock Improved techniques for training score-based generative models.
\newblock {\em Advances in Neural Information Processing Systems}, 33:12438--12448, 2020.

\bibitem[\protect\citeauthoryear{Song \bgroup \em et al.\egroup }{2020}]{song2020score}
Yang Song, Jascha Sohl-Dickstein, Diederik~P Kingma, Abhishek Kumar, Stefano Ermon, and Ben Poole.
\newblock Score-based generative modeling through stochastic differential equations.
\newblock {\em arXiv preprint arXiv:2011.13456}, 2020.

\bibitem[\protect\citeauthoryear{Song \bgroup \em et al.\egroup }{2021}]{song2021maximum}
Yang Song, Conor Durkan, Iain Murray, and Stefano Ermon.
\newblock Maximum likelihood training of score-based diffusion models.
\newblock {\em Advances in Neural Information Processing Systems}, 34:1415--1428, 2021.

\bibitem[\protect\citeauthoryear{Sung \bgroup \em et al.\egroup }{2023}]{sung2023neuroevolution}
Nicholas Sung, Jian~Cheng Wong, Chin~Chun Ooi, Abhishek Gupta, Pao-Hsiung Chiu, and Yew-Soon Ong.
\newblock Neuroevolution of physics-informed neural nets: Benchmark problems and comparative results.
\newblock In {\em Proceedings of the Companion Conference on Genetic and Evolutionary Computation}, pages 2144--2151, 2023.

\bibitem[\protect\citeauthoryear{Szymanski \bgroup \em et al.\egroup }{2023}]{szymanski2023autonomous}
Nathan~J Szymanski, Bernardus Rendy, Yuxing Fei, Rishi~E Kumar, Tanjin He, David Milsted, Matthew~J McDermott, Max Gallant, Ekin~Dogus Cubuk, Amil Merchant, et~al.
\newblock An autonomous laboratory for the accelerated synthesis of novel materials.
\newblock {\em Nature}, 624(7990):86--91, 2023.

\bibitem[\protect\citeauthoryear{Tan \bgroup \em et al.\egroup }{2025}]{tan2025fast}
Kim~Yong Tan, Yueming Lyu, Ivor Tsang, and Yew-Soon Ong.
\newblock Fast direct: Query-efficient online black-box guidance for diffusion-model target generation.
\newblock In {\em International Conference on Learning Representations}, 2025.

\bibitem[\protect\citeauthoryear{Valiant}{2009}]{valiant2009evolvability}
Leslie~G Valiant.
\newblock Evolvability.
\newblock {\em Journal of the ACM}, 56(1):1--21, 2009.

\bibitem[\protect\citeauthoryear{Wallace \bgroup \em et al.\egroup }{2023}]{wallace2023end}
Bram Wallace, Akash Gokul, Stefano Ermon, and Nikhil Naik.
\newblock End-to-end diffusion latent optimization improves classifier guidance.
\newblock In {\em Proceedings of the IEEE/CVF International Conference on Computer Vision}, pages 7280--7290, 2023.

\bibitem[\protect\citeauthoryear{Wei \bgroup \em et al.\egroup }{2023}]{wei2023select}
Zhao Wei, Jian~Cheng Wong, Nicholas Sung, Abhishek Gupta, Chin~Chun Ooi, Pao-Hsiung Chiu, My~Ha Dao, and Yew-Soon Ong.
\newblock How to select physics-informed neural networks in the absence of ground truth: A pareto front-based strategy.
\newblock In {\em 1st Workshop on the Synergy of Scientific and Machine Learning Modeling@ ICML}, 2023.

\bibitem[\protect\citeauthoryear{Wierstra \bgroup \em et al.\egroup }{2014}]{wierstra2014natural}
Daan Wierstra, Tom Schaul, Tobias Glasmachers, Yi~Sun, Jan Peters, and J{\"u}rgen Schmidhuber.
\newblock Natural evolution strategies.
\newblock {\em The Journal of Machine Learning Research}, 15(1):949--980, 2014.

\bibitem[\protect\citeauthoryear{Wong \bgroup \em et al.\egroup }{2024}]{wong2024llm2fea}
Melvin Wong, Jiao Liu, Thiago Rios, Stefan Menzel, and Yew~Soon Ong.
\newblock {LLM2FEA}: Discover novel designs with generative evolutionary multitasking.
\newblock {\em arXiv preprint arXiv:2406.14917}, 2024.

\bibitem[\protect\citeauthoryear{Xu \bgroup \em et al.\egroup }{2025}]{xulooks}
Qingshan Xu, Jiao Liu, Melvin Wong, Caishun Chen, and Yew-Soon Ong.
\newblock Looks great, functions better: Physics compliance text-to-3{D} shape generation.
\newblock In {\em International Joint Conference on Neural Networks}, 2025.

\bibitem[\protect\citeauthoryear{Yang \bgroup \em et al.\egroup }{2022}]{yang2022physics}
Zhong Liang~Ou Yang, Yang Jiang, Jian~Cheng Wong, Pao-Hsiung Chiu, Weijiang Zhao, My~Ha Dao, and Chin~Chun Ooi.
\newblock Physics compliance as a metric for neural network uncertainty.
\newblock In {\em IEEE Symposium Series on Computational Intelligence}, pages 1--7, 2022.

\bibitem[\protect\citeauthoryear{Yang \bgroup \em et al.\egroup }{2023}]{yang2023diffusion}
Ling Yang, Zhilong Zhang, Yang Song, Shenda Hong, Runsheng Xu, Yue Zhao, Wentao Zhang, Bin Cui, and Ming-Hsuan Yang.
\newblock Diffusion models: A comprehensive survey of methods and applications.
\newblock {\em ACM Computing Surveys}, 56(4):1--39, 2023.

\bibitem[\protect\citeauthoryear{Yuan \bgroup \em et al.\egroup }{2023}]{yuan2023physdiff}
Ye~Yuan, Jiaming Song, Umar Iqbal, Arash Vahdat, and Jan Kautz.
\newblock {PhysDiff}: Physics-guided human motion diffusion model.
\newblock In {\em Proceedings of the IEEE/CVF International Conference on Computer Vision}, pages 16010--16021, 2023.

\bibitem[\protect\citeauthoryear{Zeni \bgroup \em et al.\egroup }{2025}]{zeni2025generative}
Claudio Zeni, Robert Pinsler, Daniel Z{\"u}gner, Andrew Fowler, Matthew Horton, Xiang Fu, Zilong Wang, Aliaksandra Shysheya, Jonathan Crabb{\'e}, Shoko Ueda, et~al.
\newblock A generative model for inorganic materials design.
\newblock {\em Nature}, pages 1--3, 2025.

\end{thebibliography}

\end{document}